\documentclass[letterpaper, 10 pt, conference]{ieeeconf}

\IEEEoverridecommandlockouts
\usepackage{cite}
\usepackage{float}
\usepackage{graphicx}
\usepackage{amssymb}

\usepackage{amsmath}
\usepackage{mathtools}  

\usepackage{nccmath}

\usepackage{amsthm}
\usepackage[usenames,dvipsnames,svgnames,table]{xcolor}
\theoremstyle{plain}

\newtheorem{problem}{Problem}

\usepackage{float}
\floatstyle{plaintop}
\restylefloat{table}
\usepackage{MnSymbol}%
\usepackage{wasysym}%
\usepackage{caption}
\usepackage{subcaption}
\usepackage{graphicx}
\usepackage{import}

\usepackage[hidelinks]{hyperref}

\usepackage[colorinlistoftodos]{todonotes}

\usepackage{algorithm}
\usepackage{algpseudocode}

\definecolor{deepblue}{rgb}{0,0,1}

\newcommand{\correction}[1]{{#1}}

\newcommand{\SB}[1]{{{#1}}}

\begin{document}

\title{ 
\bf 
Forming and Controlling Hitches in Midair Using Aerial Robots 
}

\author{Diego S. D'Antonio, Subhrajit Bhattacharya, and David Salda\~{n}a
\thanks{$^1$The authors are with the Autonomous and Intelligent Robotics Laboratory --AIRLab-- at Lehigh University, Bethlehem, PA, 18015, USA. \texttt{Email:\{diego.s.dantonio,~sub216,~saldana\}@lehigh.edu}}
\thanks{\SB{$^2$Coauthor Bhattacharya acknowledges the support from National Science Foundation Grant No. CCF-2144246.}}
}


\maketitle

\begin{abstract}
The use of cables for aerial manipulation has shown to be a lightweight and versatile way to interact with objects. However, fastening objects using cables is still a challenge and human \correction{is required}. In this work, we propose a novel \correction{way to secure objects using hitches. The} hitch can be formed and \correction{morphed} in midair using a team of aerial robots with cables. The hitch\correction{'s shape is modeled as} a convex polygon, making it versatile and adaptable to a wide variety of objects. 
We propose an algorithm to form the hitch systematically. The steps can run in parallel, allowing hitches with a large number of robots to be formed in constant time. We develop a set of actions that include different \correction{actions} to change the shape of the hitch. We demonstrate our methods using a team of aerial robots via simulation and actual experiments.

%
%
%
%
%
%
%
%




\end{abstract}


\IEEEpeerreviewmaketitle



\section{introduction}



From ancient times, humans have been familiar with the use of ropes to \correction{secure and transport} objects. There is an old trace indicating that neanderthals used twisted fiber to tie objects up \cite{Hardy2020}. So ropes have been used even before the invention of the wheel, and nowadays, we can see ropes, cables, and strings everywhere in all types of applications. 
 Although humans have widely used them, their use in robotics has been very limited due to their high complexity.
A cable has infinite possible shapes, also known as configurations, that \correction{offers high} versatility\correction{, but} at the same time, this complicates its analysis and computation.

In aerial manipulation, external mechanisms, such as lightweight grippers \cite{mellinger2011design,pounds2011grasping, Hingston2020} and robot arms \cite{ aerialmanipulation2018-ollero, kim2013aerial, huber2013first, bellicoso2015design}, have been added to interact with objects and the environment. However, the attachment of an external mechanism on an aerial vehicle increases its system complexity \cite{keemink2012mechanical}, changing inertia, the center of mass, and overall weight.  In contrast to those types of mechanisms, ropes are lightweight and low-cost. 
%
 %
%
%
%
 %

%
%
%
%
The use of cables in aerial manipulation has existed for more than a decade, and there are significant contributions to the state of the art~\cite{Meng2022, ollero2022}. Specifically, cables attached to quadrotors have become part of science development due to their abilities and versatility. For instance, a  quadrotor is constrained to its maximum thrust in object transportation, but multiple quadrotors with cables can combine forces and increase their \correction{actual capacity} \cite{RSS2013Koushil}.
%
%
%
%
%
%
%
%
%
%
Suspended load transportation was studied with a
single cable and multiple cables \cite{sreenath2013trajectory, cruz2017cable, Cardona2019, Sanalitro2020, Kotaru2018}.
%
Although there is a significant amount of existing research in suspended load transportation,  most approaches assume that the connection between the quadrotor and the load is made in a previous stage. That is a bottleneck in autonomous transportation because it requires human intervention.

Humans \correction{interlace cables to form} hitches and knots in their daily life for multiple purposes, especially to tie, hold, and carry objects. Cowboys use hitches to \correction{tied} horses or hold objects with multiple interconnected ropes 
\cite{budworth2000}. A wide classification of hitches can be found in \cite{Bayman1977}. 
%
%
We found interesting physical hitches, such as a single diamond and a Marline hitch. Both used multiple cable intersections to hold objects.
In this manuscript, we do not make any distinction between ropes and cables. The seminal work that introduced aerial robots weaving multiple cables to create hitches was presented in~\cite{Augugliaro2013}. 
The authors focused on forming tensile structures such as bridges.


%

%
%

%

%
%


\begin{figure}[t]
\centering
\hfill
\begin{subfigure}{0.45\linewidth}
  \centering
  \includegraphics[width=\textwidth]{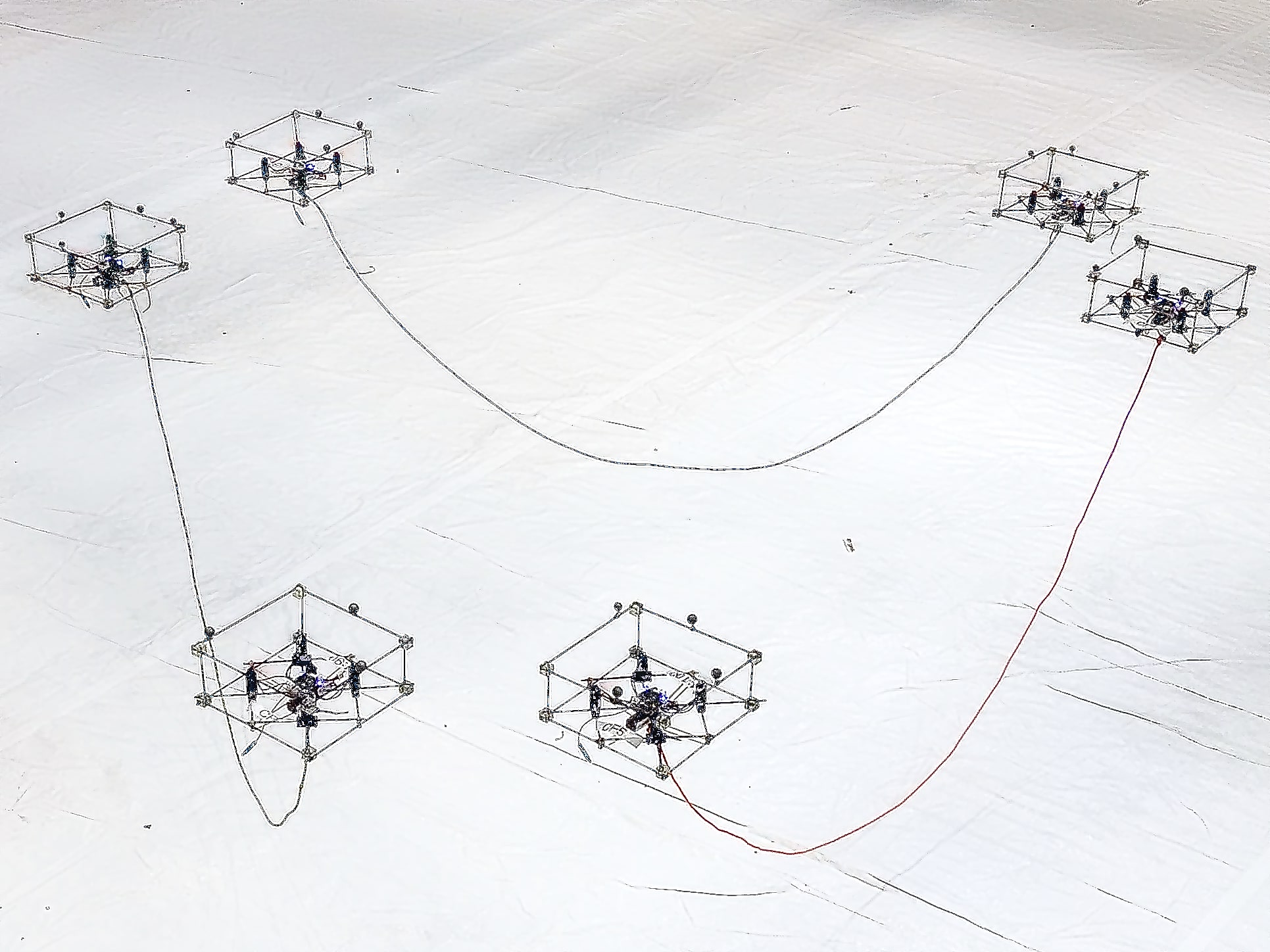}
  \caption{Flying catenary robots}
  \label{fig:exp_braid_a}
\end{subfigure}
\hfill
\begin{subfigure}{0.45\linewidth}
  \centering
  \includegraphics[width=\textwidth]{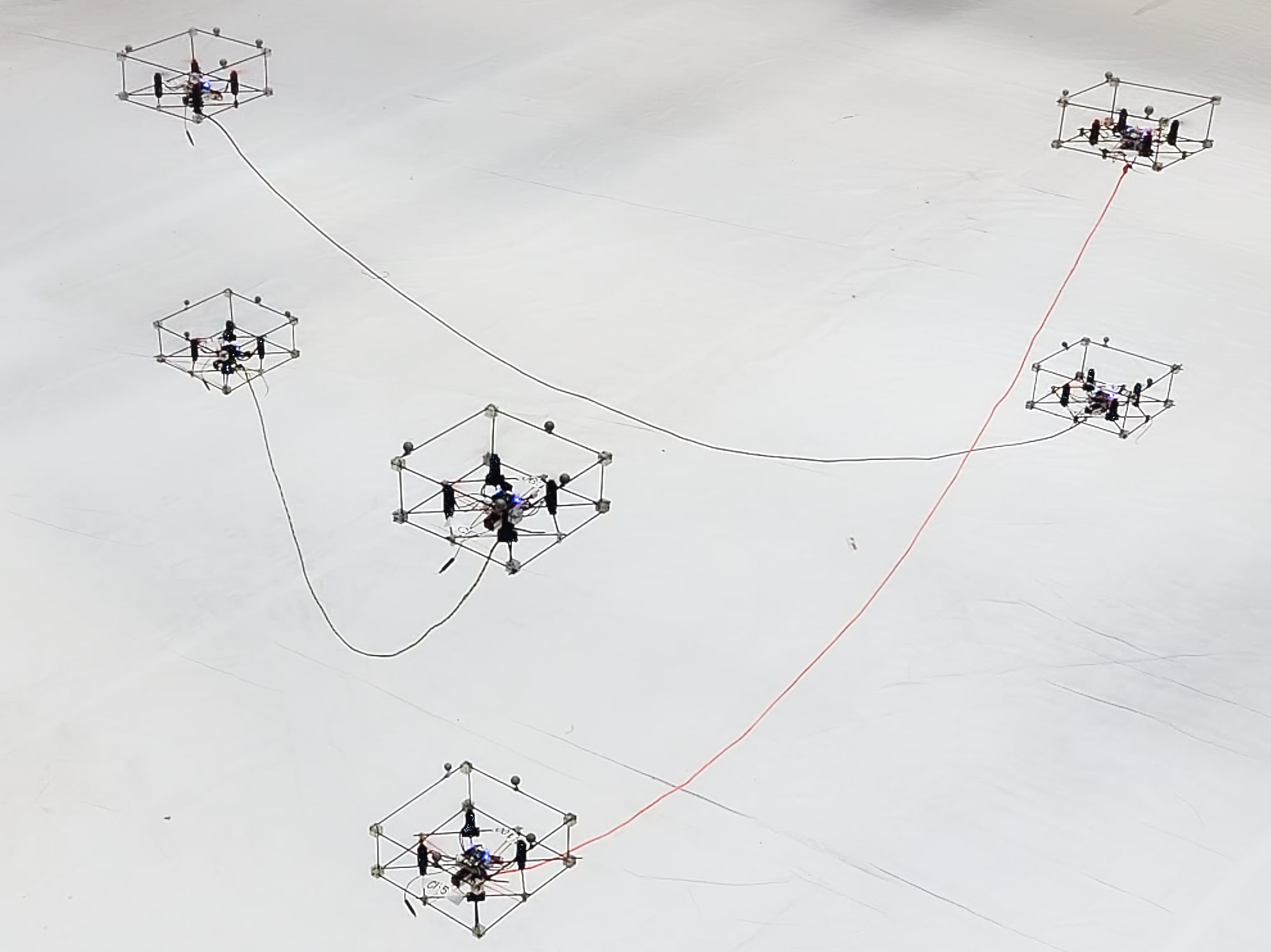}
  \caption{Interlacing cables}
  \label{fig:exp_braid_b}
\end{subfigure}
\hfill

\hfill
\begin{subfigure}{0.45\linewidth}
  \centering
  \includegraphics[width=\textwidth,trim=0cm 1cm 0cm 1.5cm,clip]{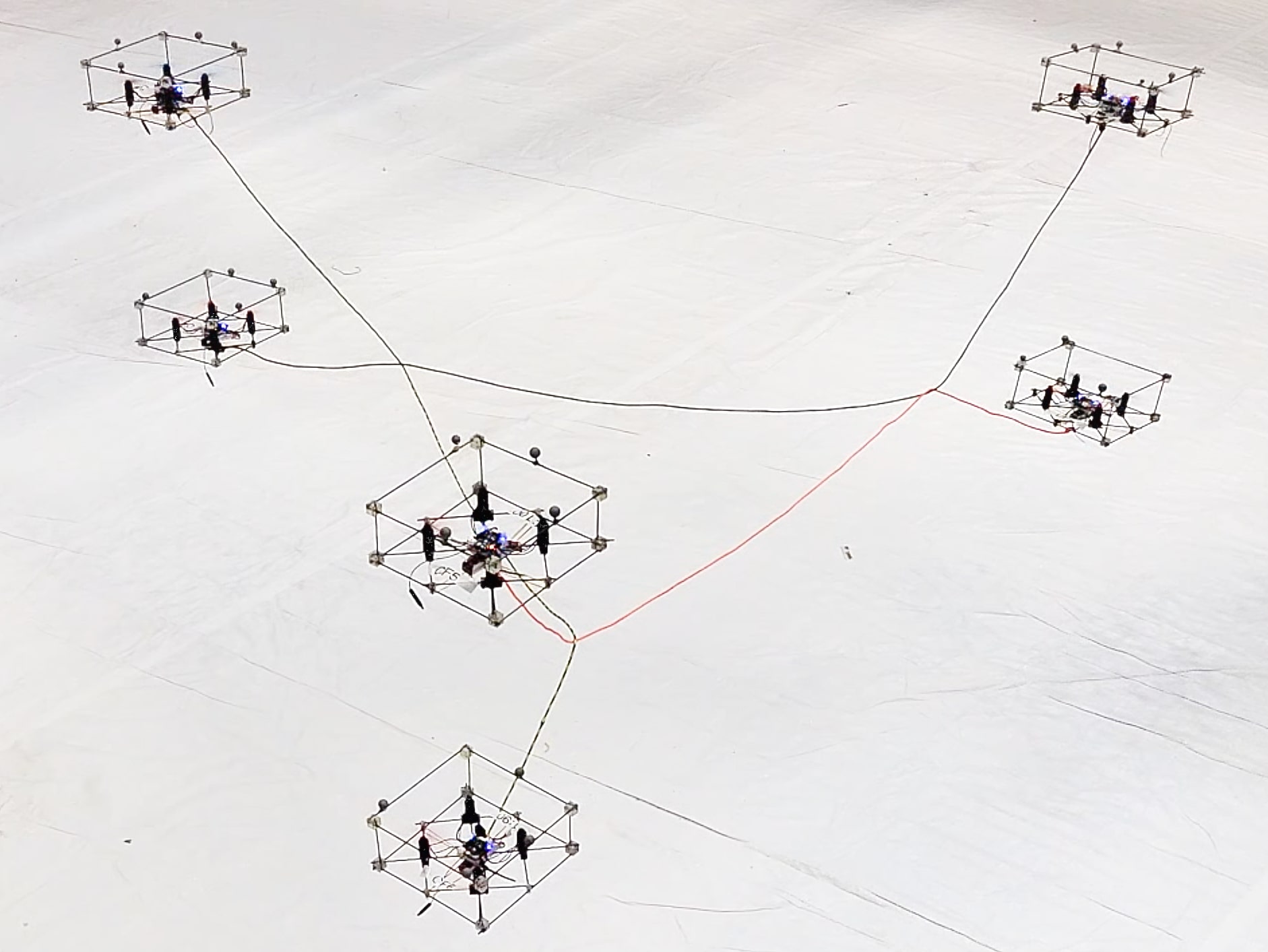}
  \caption{Forming vertices}
  \label{fig:exp_braid_d}
\end{subfigure}
\hfill
\begin{subfigure}{0.45\linewidth}
  \centering
  \includegraphics[width=\textwidth,trim=0.5cm 0cm 0.2cm 0cm,clip]{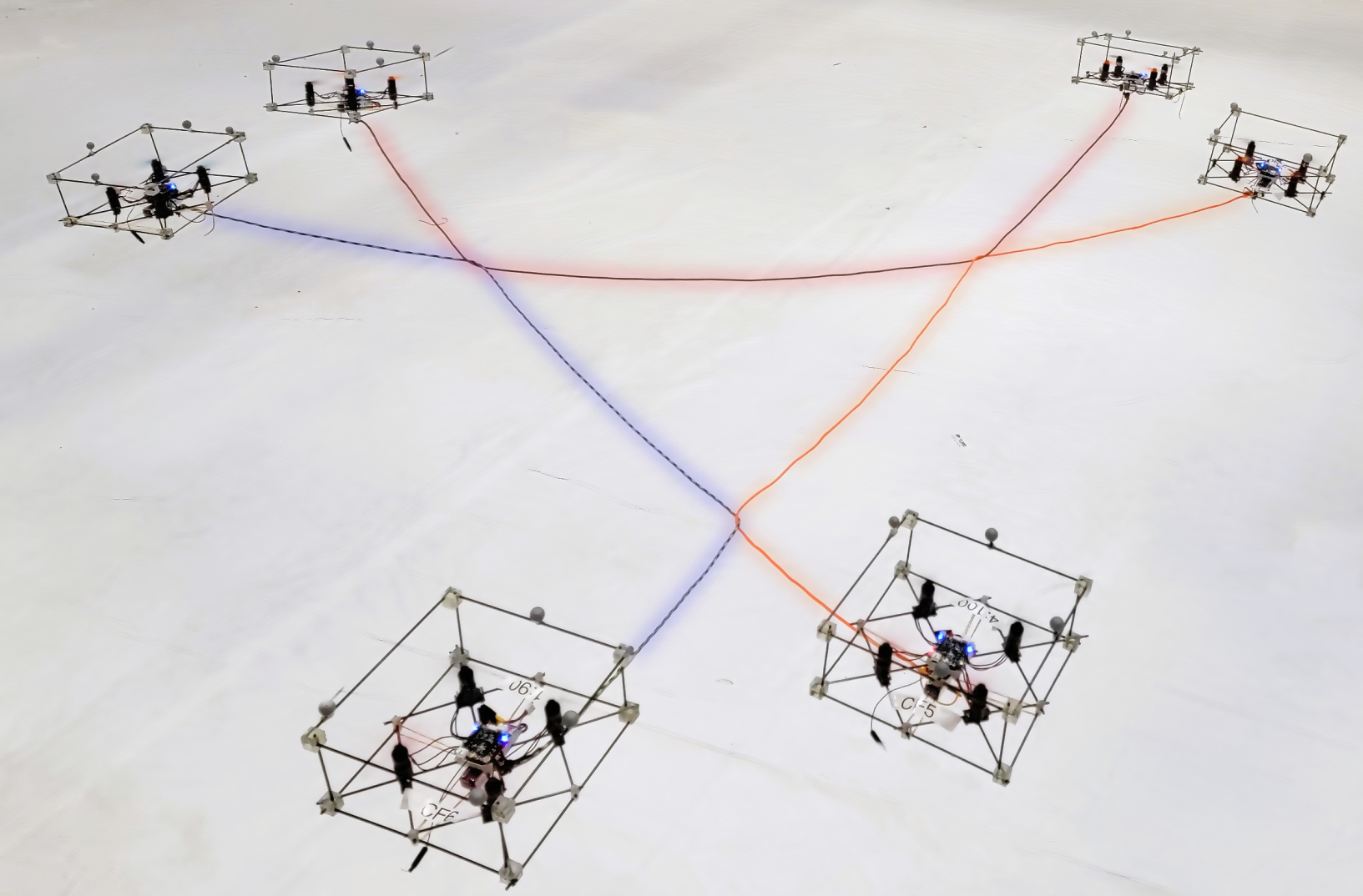}
  \caption{Triangular hitch}
  \label{fig:exp_braid_e}
\end{subfigure}
\hfill

    \caption{\correction{Six quadrotors forming} a \correction{triangular} polygonal \correction{hitch}. Video available at: 
    \url{https://youtu.be/gBVJPY7ilzc}}
    \label{fig:hitch}
\end{figure}



\begin{figure}[t]
    \centering
\begin{subfigure}{.3\linewidth}
  \centering
  \def\svgwidth{1\linewidth}
  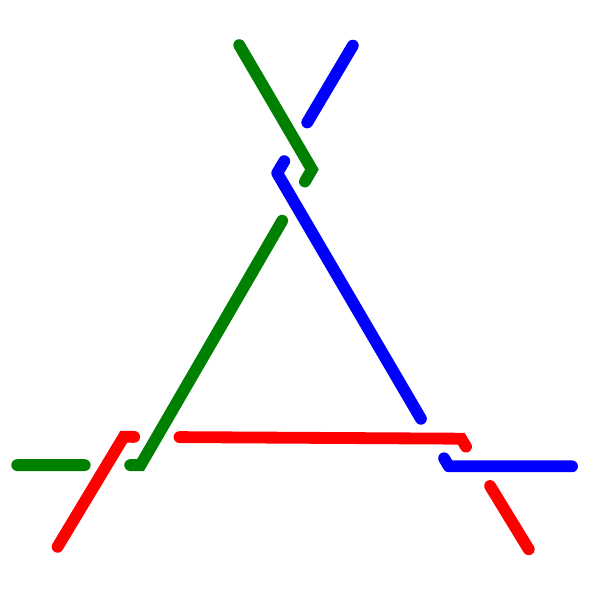
  \caption{Triangular hitch.}
  \label{fig:three_intersections}
\end{subfigure}
\hfill
\begin{subfigure}{.3\linewidth}
  \centering
  \def\svgwidth{1\linewidth}
  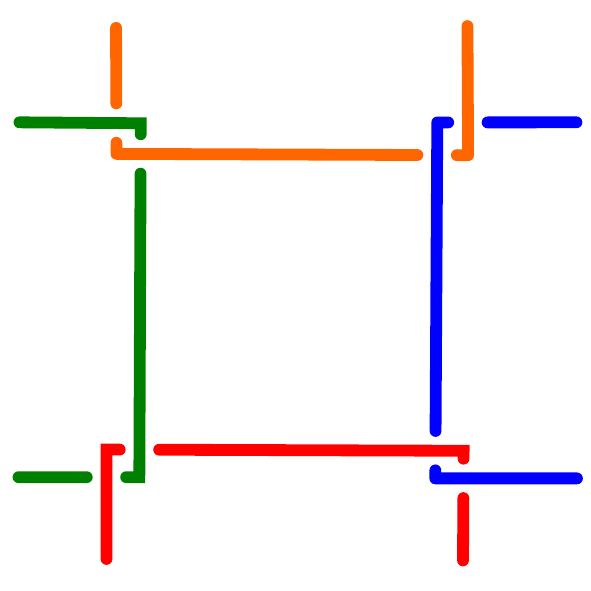
  \caption{Square hitch.}
  \label{fig:four_intersections}
\end{subfigure}
\hfill
\begin{subfigure}{.3\linewidth}
  \centering
  \def\svgwidth{1\linewidth}
  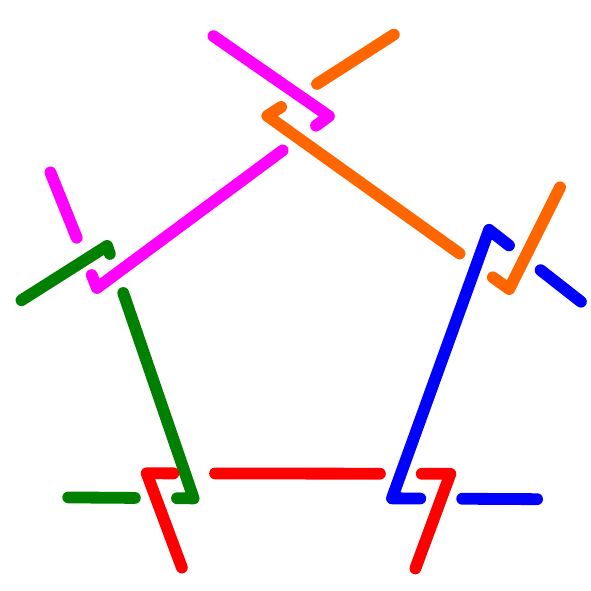
  \caption{Pentagon hitch.}
  \label{fig:five_intersections}
\end{subfigure}
\hfill
\caption{Polygonal hitches.}
\label{fig:hitches_examples}
\end{figure}

In our previous work, we introduced the catenary robot, a pair of quadrotors that control a hanging cable that describes a catenary curve.
This vehicle is used for non-prehensile manipulation with hook-shape objects~\cite{catenaryrobot} and cuboid objects~\cite{GustavoCatenary, GustavoCatenaryAdaptive}.
\correction{Since some objects require fastening for transportation purposes}, we developed an algorithm to create knots in midair~\cite{Diegoknots}, 
\correction{While tying a knot can effectively secure the object, the autonomous knot release is still difficult. Consequently, a fully autonomous transportation system using cables remains an open area of research.}
%
%
%
%
%
%
%
%
In this work, we propose a novel type of hitch and a set of actions to form it and \correction{morph} its shape \correction{in midair} using multiple catenary robots (see Fig.~\ref{fig:hitch}). The hitch is defined by a polygon, making it versatile and adaptable to a wide variety of objects. Depending on \correction{cross-sectional object shape, we can choose a convex polygon} (\correction{some polygonal} hitches are illustrated in Fig.~\ref{fig:hitches_examples}). 
\correction{For instance, a box can be transported with a square hitch}. 

%
%
The main contribution of this paper is twofold.
\textit{i)}  We introduce and formalize a new type of hitch \correction{that can be formed using aerial robots}.
\correction{The maximum size of the hitch is scalable by increasing the number of catenary robots that form it.}
\textit{ii)} We propose a new set of actions that include an algorithm to form the hitch, and change its shape.
\correction{our algorithm to form the hitch can run in parallel, allowing hitches with a large number of robots to be formed in constant time. Additionally,} Our solution is computationally efficient and runs \correction{in actual robots}. 













\section{Problem statement}

\label{section: Problem Statement}

\begin{figure}[b]
    \centering
\begin{subfigure}{0.32\linewidth}
  \centering
  \includegraphics[height=0.9\textwidth,trim=5cm 0cm 1.5cm 0cm,clip]{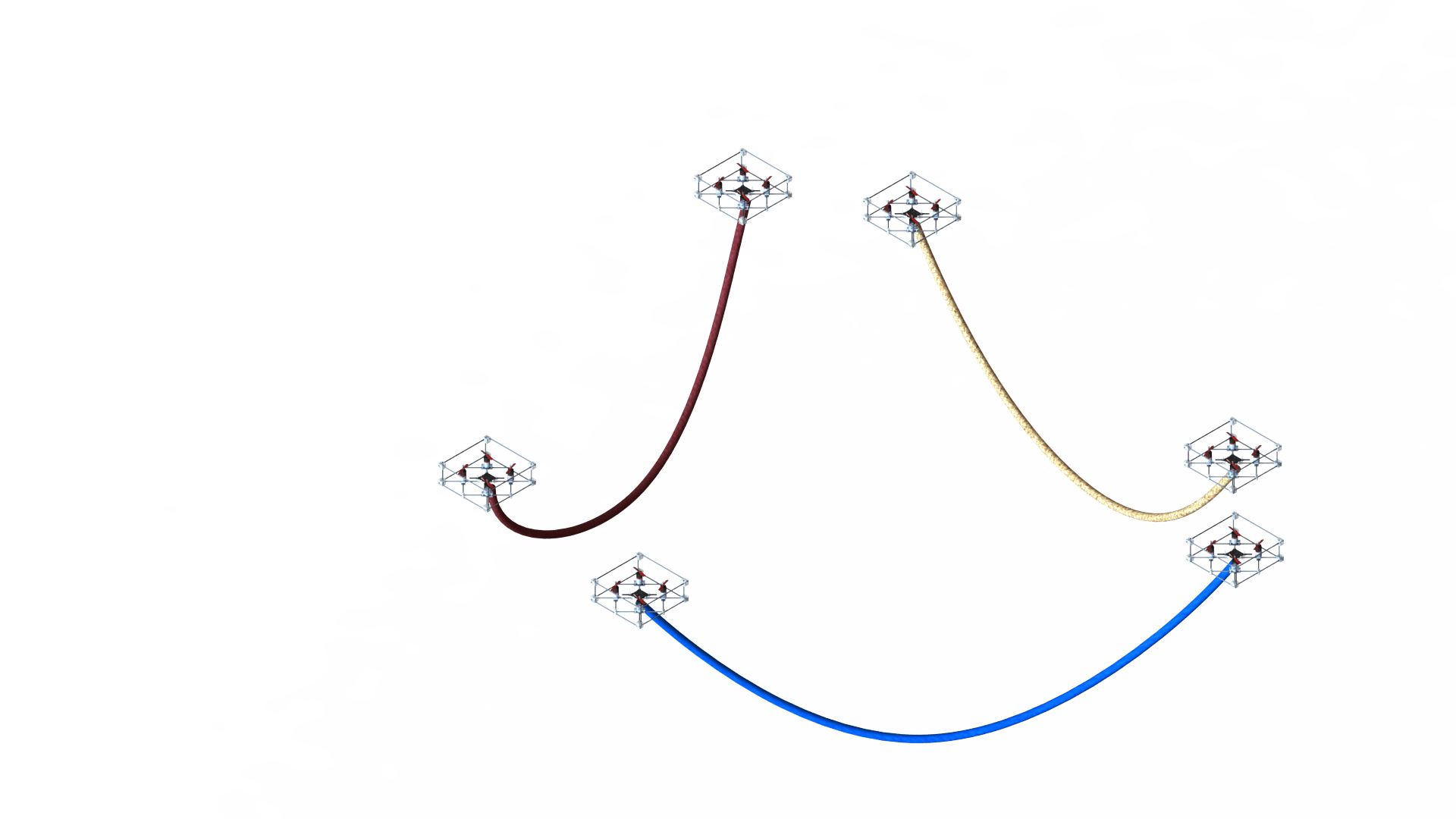}
  \caption{}
  \label{fig:stage_catenary}
\end{subfigure}
\hfill
\begin{subfigure}{0.32\linewidth}
  \centering
  \includegraphics[height=0.9\textwidth,trim=5cm 0cm 1.5cm 0cm,clip]{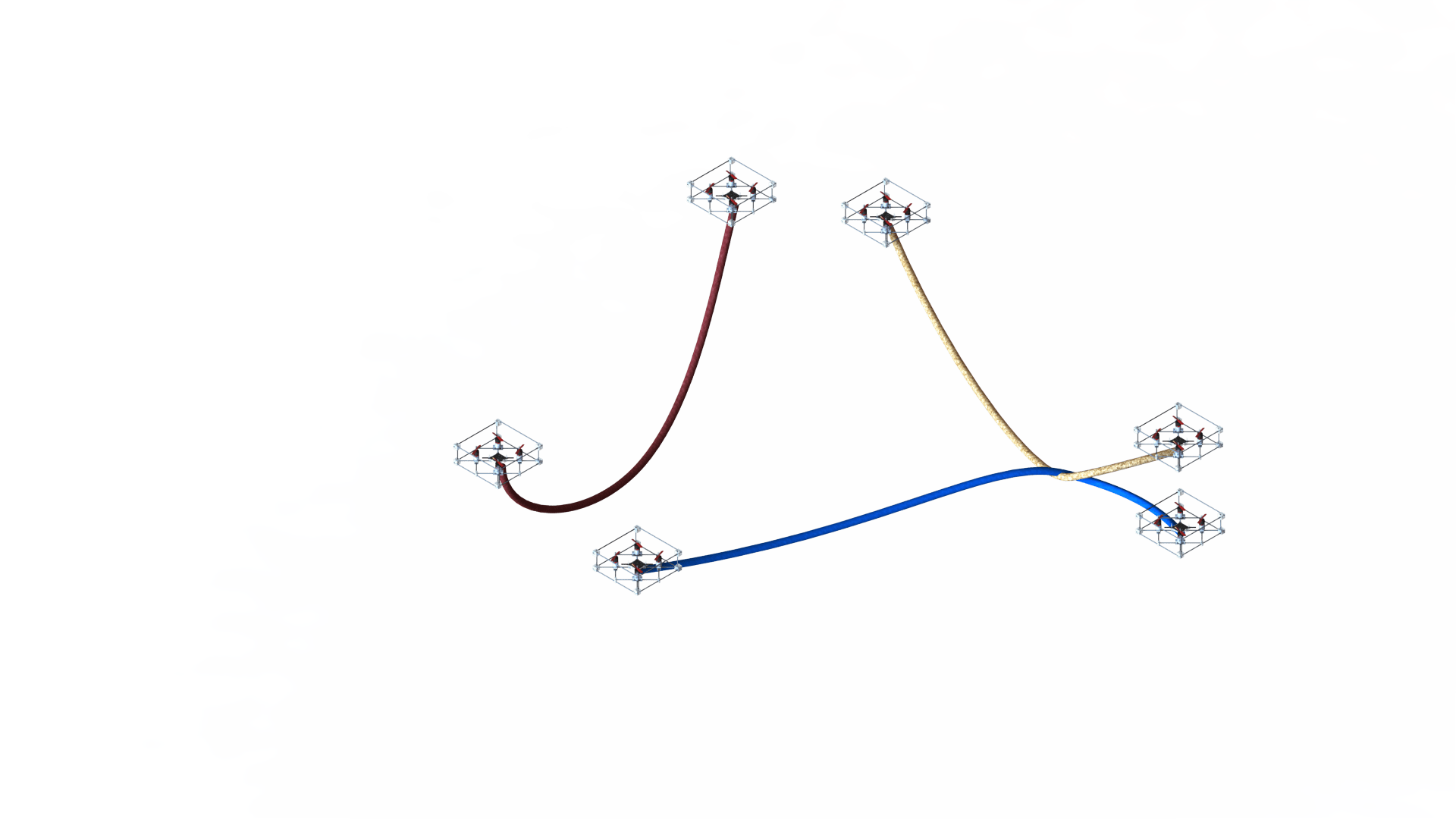}
  \caption{}
  \label{fig:stage_hitch}
\end{subfigure}
\hfill
\begin{subfigure}{0.32\linewidth}
  \centering
  \includegraphics[height=0.9\textwidth,trim=5cm 0cm 0.5cm 0cm,clip]{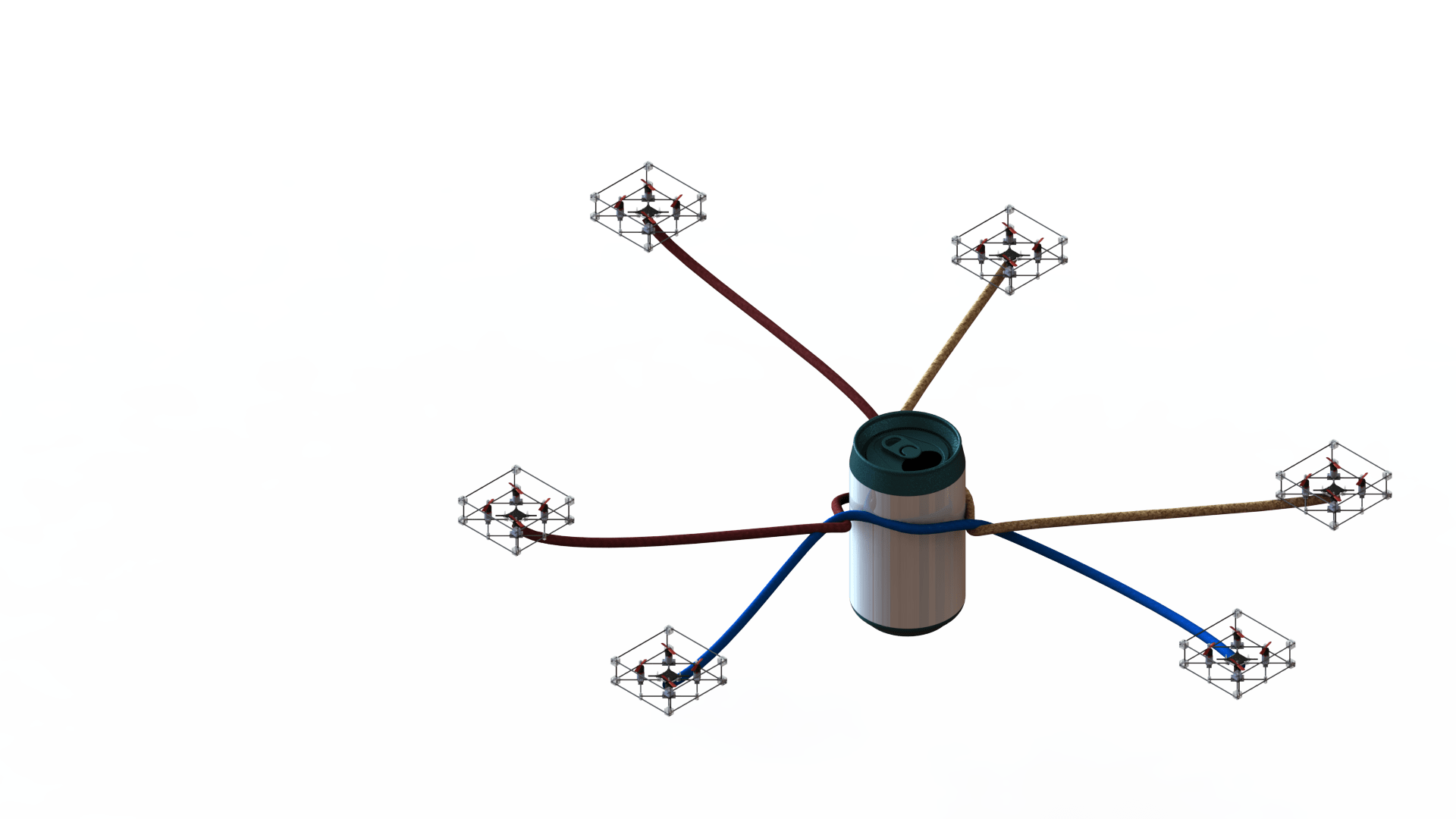}
  \caption{}
  \label{fig:stage_lift}
\end{subfigure}
\hfill
\caption{Stages of forming a hitch with aerial robots, (a) Free catenary robots, (b) Weaving a cable, and (c) Control the hitch shape.}
\label{fig:hitches_stages}
\end{figure}

Consider a team of \emph{catenary robots} \cite{catenaryrobot}, where each robot
is composed of two quadcopters attached to the ends of a flexible non-stretchable cable (see Fig.~\ref{fig:stage_catenary}).
The catenary robots can interlace their cables by passing along each other (see Fig.~\ref{fig:stage_hitch}). 
The interaction of multiple catenary robots forming a hitch creates a manipulation tool for aerial robots that is lightweight, versatile, and adjustable
(see Fig.~\ref{fig:stage_lift}).
We define a world frame in $\mathbb{R}^3$, denoted by $\{\mathcal{W}\}$, which is fixed, and its $z$-axis points upwards. 
Although our analysis uses a plane as a workspace, notice that the robots can freely move and reach the planar workspace that can be projected in the three-dimensional space.

\begin{figure}[t]
    \centering
    \includegraphics[width=0.5\columnwidth]{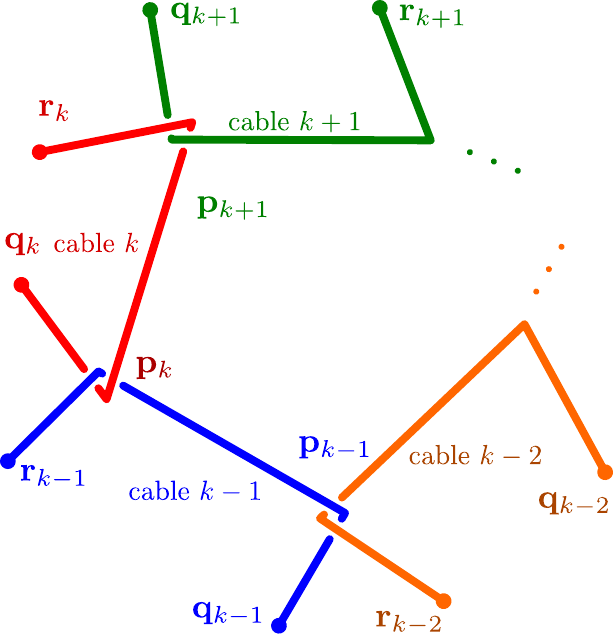}
    \caption{Notations involved in describing a general polygonal-hitch.}
    \label{fig:hitch-notations}
\end{figure}

\textbf{Polygonal-hitch: }
Hitches have been studied for many years \cite{Bayman1977, Krauel2005, Maddocks1987} and there are several types,
but in this paper, we focus on a specific type of hitch that
\correction{can be formed  usign aerial robots,}
we call \textit{it polygonal hitch}.
Based on a convex polygon, defined by a set of $n$ vertices on the Euclidean plane, i.e., $\mathcal{P} = \{\boldsymbol{p}_k\in\mathbb{R}^2, \: k = 1, \dots,n \}$,
we want to use $n$ catenary robots to \correction{interlace} their cables and form \correction{a hitch with the shape of} the polygon $\mathcal{P}$.
The vertices in $\mathcal{P}$ are numbered in an increasing order, following a clockwise order (see Fig.~\ref{fig:hitch-notations}). 
The $k$th edge of the polygon $\mathcal{P}$ goes from vertex $\boldsymbol{p}_k$ to vertex~$\boldsymbol{p}_{k+1}$.
%
The cable of each catenary robot is used to form an edge of the polygon, and therefore, we enumerate the robots accordingly to the edge where they belong.
Assuming that each end of the cable is attached to the center of mass of a quadrotor, we abstract the two quadrotors of a catenary robot as points, denoted by $\mathbf{q}_k$ and $\mathbf{r}_k$.
In this way, the points $(\mathbf{q}_k, \,\mathbf{r}_k)$ also represent the end points of the cable.
The length of the cable is $L_k$ and we assume that the polygon satisfies $\| \mathbf{p}_{k+1} - \mathbf{p}_{k} \|< L_k$.
Notice that vertex $\mathbf{p}_k$ has two adjacent quadrotors located at  $\mathbf{r}_{k-1}$ and $\mathbf{q}_k$.
Due to the cyclic nature of the system, any index greater than $n$ will be reduced modulo $n$ (plus $1$ in order to make the indices start at $1$ rather than at~$0$). Thus, for notational convenience, whenever we refer to an index $k+1$, we mean the index $(k \mod n) + 1$. 
%
%
%

\textbf{Hitch configuration: }
The configuration of a polygonal-hitch is determined by the location of the quadrotors $(\mathbf{q}_k,\mathbf{r}_k)$, and the vertices of the polygon $\mathbf{p}_k$ for $k=1,\dots,n$,
denoted by the set 
$$\mathcal{C} = \{(\mathbf{q}_k,\mathbf{r}_k,\mathbf{p}_k),\: k=1, \dots,n\}.$$
Each cable is always in tension, and can be represented by three straight lines.
Assuming negligible friction at the interaction between the cables, the \correction{uniform} tension over the entirety of each cable.
We denote the tension on the $k$-th cable by $T_k$.

%

The polygonal-hitch offers a new versatile way to interact and manipulate objects since its geometric representation allows rotation, translation and shape adaptation.
In this work, we focus on finding a class of polygonal hitches that can be formed and controlled systematically.

\begin{problem}
Given a polygon $\mathcal{P}$ with $n$ vertices and a team of~$n$ catenary robots, 
design the strategy to form a polygonal-hitch and a set of actions to be able to change its shape on the fly.
\end{problem}





\section{Polygonal-Hitches in Equilibrium}
\label{sec:hitches}
 In this section, we propose a class of polygonal-hitches that can maintain their shape in midair.
 The key is to analyze the configurations that lead the tension of the cables to an equilibrium.
 \correction{Base on static analysis, we design} a practical solution to find a configuration for a polygonal-hitch $\mathcal{C}$ \correction{for} a given polygon $\mathcal{P}$.
Examples of this class of flying hitches can be seen in Fig.~\ref{fig:hitches_examples}.




\






\subsection{Equilibrium Analysis of a Static Hitch}


\begin{figure}[b]
    \centering
    \includegraphics[width=0.25\textwidth,trim=0.cm 0cm 0.cm 0.3cm,clip]{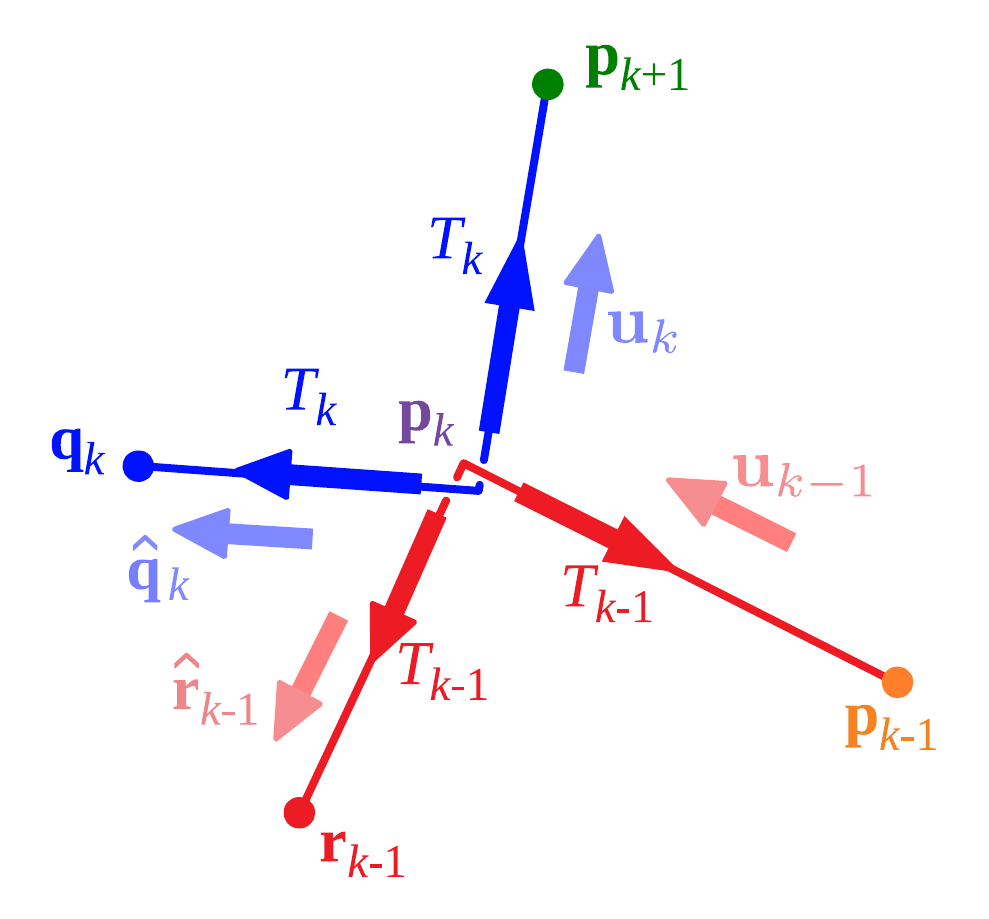}
    \caption{Tensions at the $k$-th vertex of a polygonal-hitch.}
    \label{fig:force-eqb}
\end{figure}
A vertex is formed by interlacing two cables, forming an x-like shape with four tensions (see Fig.~\ref{fig:force-eqb}).
We analyze the tensions at the  $k$-th vertex to be in force equilibrium.
The direction of the $k$th edge of the polygon, $\correction{({\mathbf{p}}_{k}, {\mathbf{p}}_{k+1})}$, is denoted by the unit vector
\begin{equation}
{\mathbf{u}}_{k} = \frac{\mathbf{p}_{k+1} - \mathbf{p}_k}{\|\mathbf{p}_{k+1} - \mathbf{p}_k\|}.
\label{eq:u}
\end{equation}
The $k$-th vertex at location $\mathbf{p}_k$ is in equilibrium when all the tensions add up to zero, meaning that
the following equation has to be satisfied,
\begin{equation}
    T_k \widehat{\mathbf{q}}_k 
    + T_k {\mathbf{u}}_{k} 
    + T_{k-1} \correction{\widehat{\mathbf{r}}_{k-1}}
    - T_{k-1} {\mathbf{u}}_{k-1} 
    = 0, \\
     \label{eq:force-general}
\end{equation}
where the unit vectors are,
$$
\widehat{\mathbf{q}}_k = \frac{\mathbf{q}_k - \mathbf{p}_k}{\|\mathbf{q}_k - \mathbf{p}_k\|},
\quad\text{and}\quad
\widehat{\mathbf{r}}_{k-1} = \frac{\mathbf{r}_{k-1} - \mathbf{p}_{k}}{\|\mathbf{r}_{k-1} - \mathbf{p}_{k+1}\|}.
$$


%
%
\noindent
In general, the whole system has $2n$ equations (there are $n$ vertices, and each vertex has to satisfy the vector equation \correction{in}~\eqref{eq:force-general} for $x$ and $y$ coordinates). 
The position of all vertices are part of the input, so the vectors ${\mathbf{u}}_{k}$ and ${\mathbf{u}}_{k-1}$ are known, but the tensions, $T_k,$ and cable orientations $(\widehat{\mathbf{q}}_k,\correction{\widehat{\mathbf{r}}_{k-1})}$ are unknown.
Since each cable orientation is unitary and can be determined by a single parameter, the total number of unknowns is $3n$.
Therefore, the system is underdetermined, and there are in general infinitely many solutions.

\subsection{A Specific Solution for Equilibrium}

In order to find a practical solution for the equilibrium equation in \eqref{eq:force-general}, we consider a special case where all the cables have the same tension~$T>0$, \emph{i.e.}, $T_1 =...= T_{n}=T$, and 
the cables are aligned with the polygon edges, \emph{i.e.}, $\widehat{\mathbf{r}}_{k-1} = -{\mathbf{u}}_{k}$ and $\widehat{\mathbf{q}}_k = {\mathbf{u}}_{k-1}$, for all $k=1,\dots,n$.
%
%
Then, we can easily verify that our specific solution 
\begin{equation}
    \widehat{\mathbf{q}}_k \!=\! {\mathbf{u}}_{k-1},~~
    \widehat{\mathbf{r}}_{k-1} \!=\! -{\mathbf{u}}_{k}
    ~~\text{and}~~ 
    T_k \!=\! T,
    \label{eq:special-hitch}
\end{equation}
satisfies the equilibrium equation in \eqref{eq:force-general} for any constant~$T~>~0$.
%
In the rest of the paper, we will \correction{focus on} this special class of solutions for simplicity.
Now that the orientation of the cables is defined, we only need to compute the location of the end points to find a configuration~$\mathcal{C}$ for a hitch in equilibrium.

\subsection{Determining Robot Positions}
\label{sec:robots_positions}
For a given polygon shape $\mathcal{P}$, 
the length of the $k$th edge is $l_k = \|\mathbf{p}_{k+1} - \mathbf{p}_k\|$.
Suppose the $k$-th cable has a total length of $L_k>l_k$.
Using this constraint, and the solutions for the unit vectors $\widehat{\mathbf{q}}_k$ and $\widehat{\mathbf{r}}_{k-1}$  in \eqref{eq:special-hitch}, we can compute appropriate positions for the robots $(\mathbf{q}_k, \mathbf{r}_k)$ as follows.

\begin{figure*}[t]
    \centering
\hfill
\begin{subfigure}{0.323\linewidth}
    \centering
    \includegraphics[width=\textwidth]{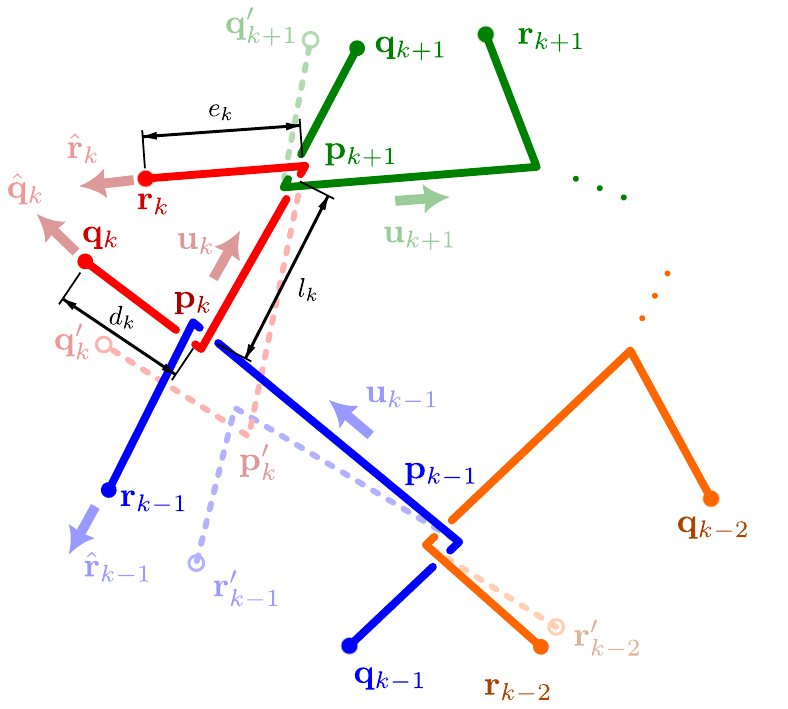}
    \caption{Action 2: Moving vertex $\mathbf{p}_k$.}
    \label{fig:hitch-special}
\end{subfigure}
\hfill
\begin{subfigure}{0.323\linewidth}
    \centering
    \includegraphics[width=\textwidth]{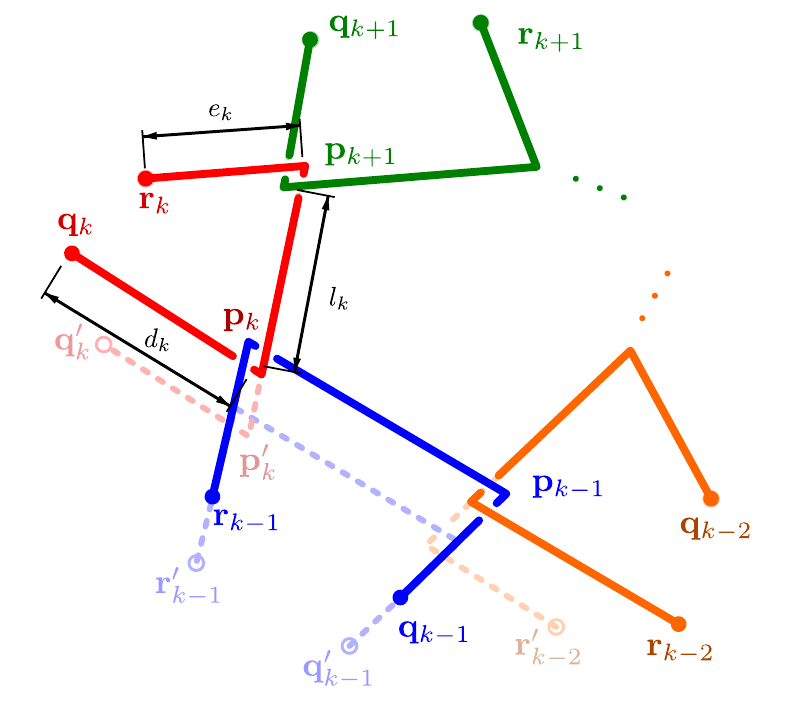}
    \caption{Action 3: Moving edge $(\mathbf{p}_{k-1}, \mathbf{p}_{k})$.}
    \label{fig:hitch-special_edge}
\end{subfigure}
\hfill
\begin{subfigure}{0.323\linewidth}
    \centering
    \includegraphics[width=\textwidth]{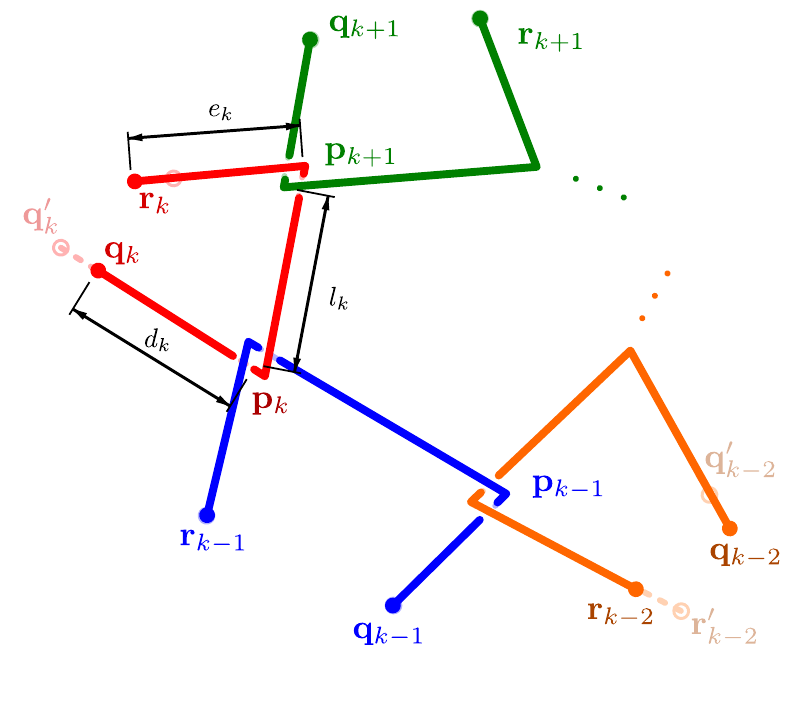}
    \caption{Action 4: Adjusting cables $k-1$ and $k$.}
    \label{fig:hitch-special_adjust}
\end{subfigure}
\hfill
    \caption{Multiple cables forming a section of a polygonal-hitch. The dashed lines represent the actions of moving a vertex, edge, and adjusting cables.}
    \label{fig:hitch2}
\end{figure*}

For each $k=1,2,\dots,n$,
choose a distance $d_k < L_k - l_k$ (or, $e_k < L_k - l_k$), at which to place the robots $\mathbf{q}_k$ from the desired polygon vertex $\mathbf{p}_k$ (or, the robot $\mathbf{r}_k$ from the desired polygon vertex $\mathbf{p}_{k+1}$), and define $e_k = L_k - l_k - d_k$ (or, $d_k = L_k - l_k - e_k$) so that $d_k + e_k + l_k = L_k$.
See Fig.~\ref{fig:hitch-special} for illustration.
Then the position vectors of the robots are given by 
\begin{eqnarray}
\mathbf{q}_k & ~=~ & \mathbf{p}_k ~+~ d_k \, {\mathbf{u}}_{k-1}, \nonumber \\
\mathbf{r}_k & ~=~ & \mathbf{p}_{k+1} ~-~ e_k \,{\mathbf{u}}_{k+1}.
\label{eq:position-equation}
\end{eqnarray}
%
\textbf{Balanced configuration: }
A special case is when the cable distances $d_k$ and $e_k$ are equal. We can compute them by,
$$
d_k=e_k=\frac{L_k - l_k}{2}.
$$
 This special case is useful to find an initial equilibrium configuration for a given polygon $\mathcal{P}$. Then, the balanced configuration is denoted by the set,
\begin{equation}
\bar{\mathcal{C}} = \{(\bar{\mathbf{q}}_k,\bar{\mathbf{r}}_k,\mathbf{p}_k),\: k=1, \dots,n\},
\label{eq:initial_c}
\end{equation}
where $\bar{\mathbf{q}}_k = \mathbf{p}_k+\frac{L_k - l_k}{2}{\mathbf{u}}_{k-1}$, and $\bar{\mathbf{r}}_k = \mathbf{p}_{k+1} \SB{-} \frac{L_k - l_k}{2}\mathbf{u}_{k+1}.$

\section{Actions for hitch manipulation}
\label{sec:actions}

In this section, we present four actions that allow robots to form a hitch and change its shape.
\correction{Assuming a} quasistatic motion. We compute robot positions and trajectories, and then use  a classical position and trajectory-tracking controller for quadrotors \cite{lee2010geometric, mellinger2011}.

\begin{figure}[b]
\centering
\begin{subfigure}{.3\linewidth}
  \centering
  \def\svgwidth{1\linewidth}
  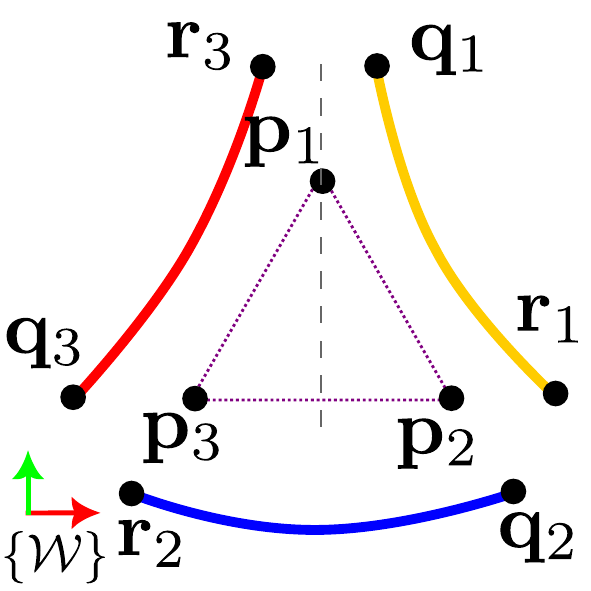
  \caption{Step 1}
  \label{fig:braid_method_catenaries}
\end{subfigure}
\hfill
\begin{subfigure}{.3\linewidth}
  \centering
  \def\svgwidth{1\linewidth}
  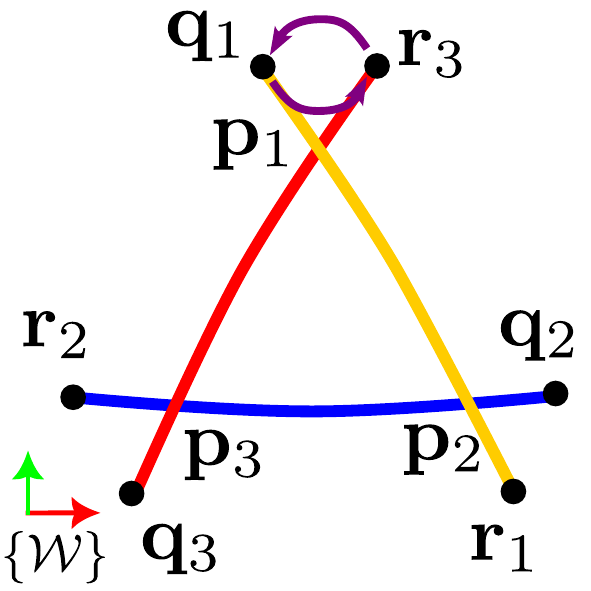
  \caption{Step 2}
  \label{fig:braid_method_first_intersection}
\end{subfigure}
\hfill
\begin{subfigure}{.3\linewidth}
  \centering
  \def\svgwidth{1\linewidth}
  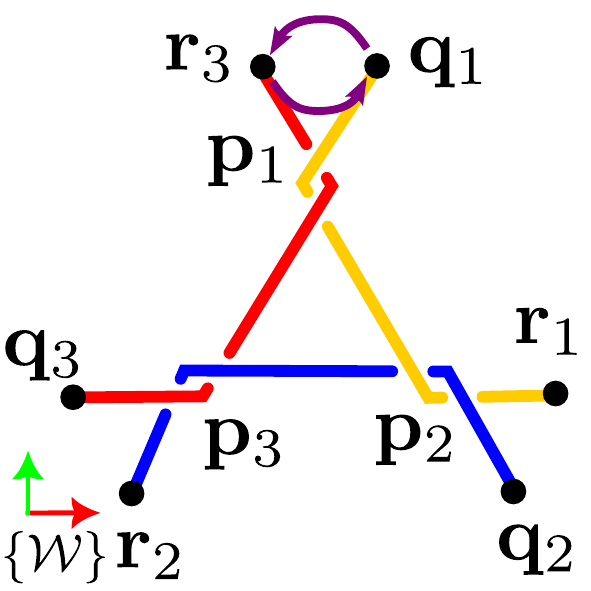
  \caption{Step 3}
  \label{fig:braid_method_second_intersection}
\end{subfigure}
\hfill
\caption{Action 1: Forming a hitch in three steps.}
\label{fig:braid_method}
\end{figure}

\subsection{Action 1: Forming a hitch}
\label{subsec:forming a hitch}

The objective of this action is to use $n$ catenary robots, initially disconnected as illustrated in Fig.~\ref{fig:braid_method_catenaries}, and interlace them to form a hitch configuration $\bar{\mathcal{C}}$:
We can form a hitch configuration~$\bar{\mathcal{C}}$ in three steps:
\begin{itemize}
    \item \textit{Step 1:} For each catenary robot $k=1,\dots,n$, move the end points of the cables, $(\mathbf{q}_k, \mathbf{r}_k)$, specified by the configuration $\bar{\mathcal{C}}$ (see Fig. \ref{fig:braid_method_catenaries}).
    \item \textit{Step 2:} Each pair of quadrotors \correction{at locations} $(\mathbf{q}_k, \mathbf{r}_{r-1})$ swap their places following a circular trajectory (or any collision-free trajectory) as illustrated in Fig.~\ref{fig:braid_method_first_intersection}.
    \item \textit{Step 3:} The same quadrotors swap their places again, \correction{interconnecting the cables and} form\correction{ing the $k$-th} vertex of the polygon~$\mathbf{p}_k$ (see Fig.~\ref{fig:braid_method_second_intersection}).
\end{itemize}
%
%
We highlight that the Step \correction{1 ,} 2, and 3 can be performed in parallel, so the time to create a polygonal-hitch of $n$ vertices is independent of $n$, taking the \correction{always constant} time.

\subsection{Action 2: Moving a vertex}

This action is focused on changing the shape of the polygon by moving a single vertex. 
We start with a configuration that forms a polygon $\mathcal{P}$ and is transformed to a polygon $\mathcal{P}'$.
Both polygons differ in a vertex $\mathbf{p}_k\in\mathcal{P}$ that we denote by $\mathbf{p}_k'\in\mathcal{P}'$.
Fig.~\ref{fig:hitch-special} illustrates the $\mathcal{P}$ in solid lines and the difference with $\mathcal{P}'$ in dashed lines.

From \eqref{eq:position-equation}, it can be noted that for a given $d_k$, 
The computation of $\mathbf{q}_k$ depends only on the position of the vertices $\mathbf{p}_k$ and $\mathbf{p}_{k-1}$. Similarly, for a given $e_k$, $\mathbf{r}_k$ depends only on the position of the vertices $\mathbf{p}_{k+1}$ and $\mathbf{p}_{k+2}$ (based on \eqref{eq:u}).
As a consequence, it is easy to check that, for a given value of $e_k$ and $d_{k-1}$, $\mathbf{p}_k$ is involved only in the expressions of $\mathbf{q}_k,\mathbf{r}_{k-1},\mathbf{q}_{k+1}$ and $\mathbf{r}_{k-2}$.
Thus, in order to change the position of a single vertex of the polygon, from $\mathbf{p}_k$ to $\mathbf{p}_k'$, while keeping the positions of all other vertices fixed, it is sufficient to recompute the end points (using \eqref{eq:position-equation}) and change the positions of the robots  $\mathbf{q}_k,\mathbf{r}_{k-1},\mathbf{q}_{k+1}$ and $\mathbf{r}_{k-2}$ only.
We can see in Fig.~\ref{fig:hitch-special} that the vertex $\mathbf{p}_k$ can be moved to $\mathbf{p}_k'$ by changing only $\mathbf{q}_k,\mathbf{r}_{k-1},\mathbf{q}_{k+1}$ and $\mathbf{r}_{k-2}$.

\subsection{Action 3: Moving an edge}
This action is focused on changing the shape of the polygon $\mathcal{P}$ by moving a single edge. 
The $k$-th edge, $(\mathbf{p}_{k}, \mathbf{p}_{k+1})$ from the polygon, $\mathcal{P}$ is moved to a new location $(\mathbf{p}_{k}', \mathbf{p}_{k+1}')$, forming a new polygon $\mathcal{P}'$.
Both polygons share the same vertices, except for vertices the edge, $(\mathbf{p}_{k}', \mathbf{p}_{k+1}')$.

Similar to Action 2, only four robots need to be moved to perform this action. In this case, we can move the edge~$k$ by translating the two endpoints of the $k$th catenary robot $\mathbf{r}_{k}$ and $\mathbf{q}_{k}$ along the directions $\widehat{\mathbf{r}}_{k}$ and $\widehat{\mathbf{q}}_{k}$ respectively. At the same time, the end points $\mathbf{r}_{k-1}$ and $\mathbf{q}_{k+1}$ need to be moved to maintain the vertices at the new locations $\mathbf{p}_{k}'$ and $\mathbf{p}_{k+1}'$.
As illustrated in Fig.~\ref{fig:hitch-special_edge}, we only need to move four quadrotors, $\correction{\mathbf{r}_{k-2}},\mathbf{q}_{k}, \mathbf{r}_{k-1}, \text{ and }\correction{\mathbf{q}_{k-1}}$, to move an edge.



\subsection{Action 4: Adjusting the cable}
In the balanced configuration $\bar{\mathcal{C}}$, the cable distances $e_k$ and $d_k$ are the same for each edge, but this property is not always maintained after applying Actions 2 and 3.
The problem is that small values of, $e_k$ and $d_k$, limit the potential changes for a new polygon $\mathcal{P}'$.
For this purpose, 
 after performing Actions 2 and 3, we adjust the cable to create a balanced configuration $\bar{\mathcal{C}}$ for the new polygon $\mathcal{P}'$.
 Fig.~\ref{fig:hitch-special_edge} illustrates a configuration where $e_k\neq d_k$, and $e_{k-2}\neq d_{k-2}$. Therefore, cables $k$ and $k-2$ need to be adjusted to achieve a balanced configuration as illustrated in \ref{fig:hitch-special_adjust}.
 
 In order to adjust cable $k$, the endpoints $\mathbf{r}_{k}$ and $\mathbf{q}_{k}$ will be moved the same distance but in opposite directions along the cable lines $\widehat{\mathbf{r}}_{k}$ and $-\widehat{\mathbf{q}}_{k}$ respectively. Therefore, the new positions for the end points after the adjustment are the same as in \eqref{eq:initial_c} $\bar{\mathbf{r}}_k'$ and $\bar{\mathbf{q}}_k'$ for the new polygon $\mathcal{P}'$.

\section{Experiments}


We evaluate each of the four actions discussed in Section \ref{sec:actions}.  
First, the catenary robots start flying and form polygonal-hitches in mid-air. 
Second, we show that we can control a simple vertex in the polygonal-hitch by manipulating the robot's position.  
Third, we demonstrate that our system can move an edge. 
Finally, we show that we can adjust the cable during the flight.
We validate our method for polygonal-hitches in simulations and actual robots \SB{(see attached multimedia video)}.

%

\subsection{Simulations}
\label{section: simulations}
\begin{figure}[t]
\centering
\includegraphics[width=0.7\linewidth]{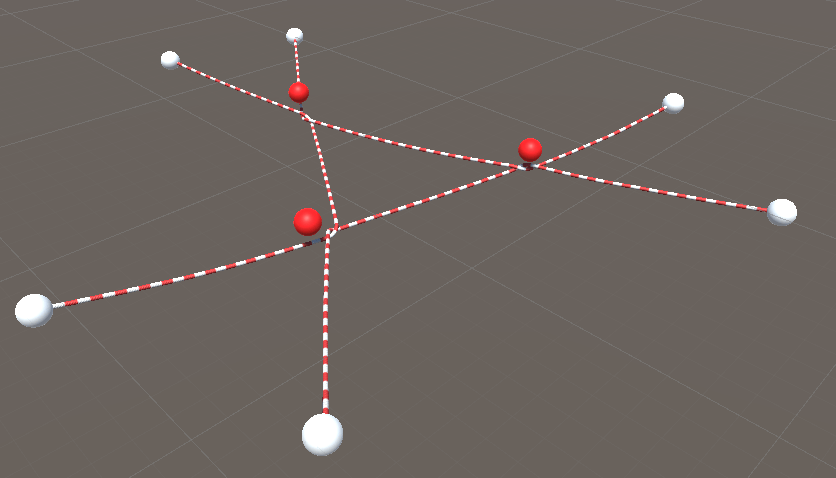}
\caption{Realistic simulation, red spheres are the desired intersections, point robots are the white color.}
\label{fig:simulation}
\end{figure}

Using a 
 realistic 3D simulator, we are able to quickly implement and test different types of maneuvers that involve cables. We performed experiments with the Obi Rope Unity package version 6.3, which is based on an advanced particle physics engine. The Obi Rope is optimized to deal with the infinite states of a rope. However, it becomes unstable once we include more than five ropes. 
We implemented a polygonal-hitch with three cables, as shown in Fig.~\ref{fig:simulation}.  
In the simulation, we are able to analyze the effect of the friction between cables, and the performance of our quasi-static approach, before running experiments with actual robots.
Our simulation framework for polygonal-hitches is open-source and publicly available\footnote{The source code for simulations and actual robots is available at\\ \url{https://github.com/swarmslab/Forming-hitches}}.

\subsection{Experiments with actual robots}
\label{section: real robots}

In our experimental testbed, we used the geometrical controller for quadrotors \cite{mellinger2011}, \correction{on the} crazyswarm \correction{framework}~\cite{crazyswarm}, which allows us to control eight robots simultaneously. 
Every quadrotor has the same components and dimensions; its weight is $132 \ g$ \correction{and, cable length of $L_{k}~=~2~m$}.
We use a single Crazyradio PA 2.4 GHz USB dongle to communicate the computer with the robots.
\correction{The localization of the quadrotors is obtained using} motion capture system (Optitrack) operating at 120 Hz for localization \SB{of the quadrotors}. 
Although we placed markers on the cables for performance analysis, our method works open-loop and only \SB{uses} the location of the quadrotors as feedback.

To obtain the intersection \SB{of the} cables,
we placed three markers in the middle of \SB{each} cable. Then, we apply a linear regression to approximate each cable section. Furthermore, we can obtain the intersections between \SB{the} lines to the desired intersections. Our evaluation metric is based on the error between the estimated intersection and the desired intersection.
In the following four experiments, validate each of the four proposed actions.


%



\begin{figure}[t]
\centering

\begin{subfigure}{0.75\linewidth}
  \centering
  \includegraphics[width=\textwidth]{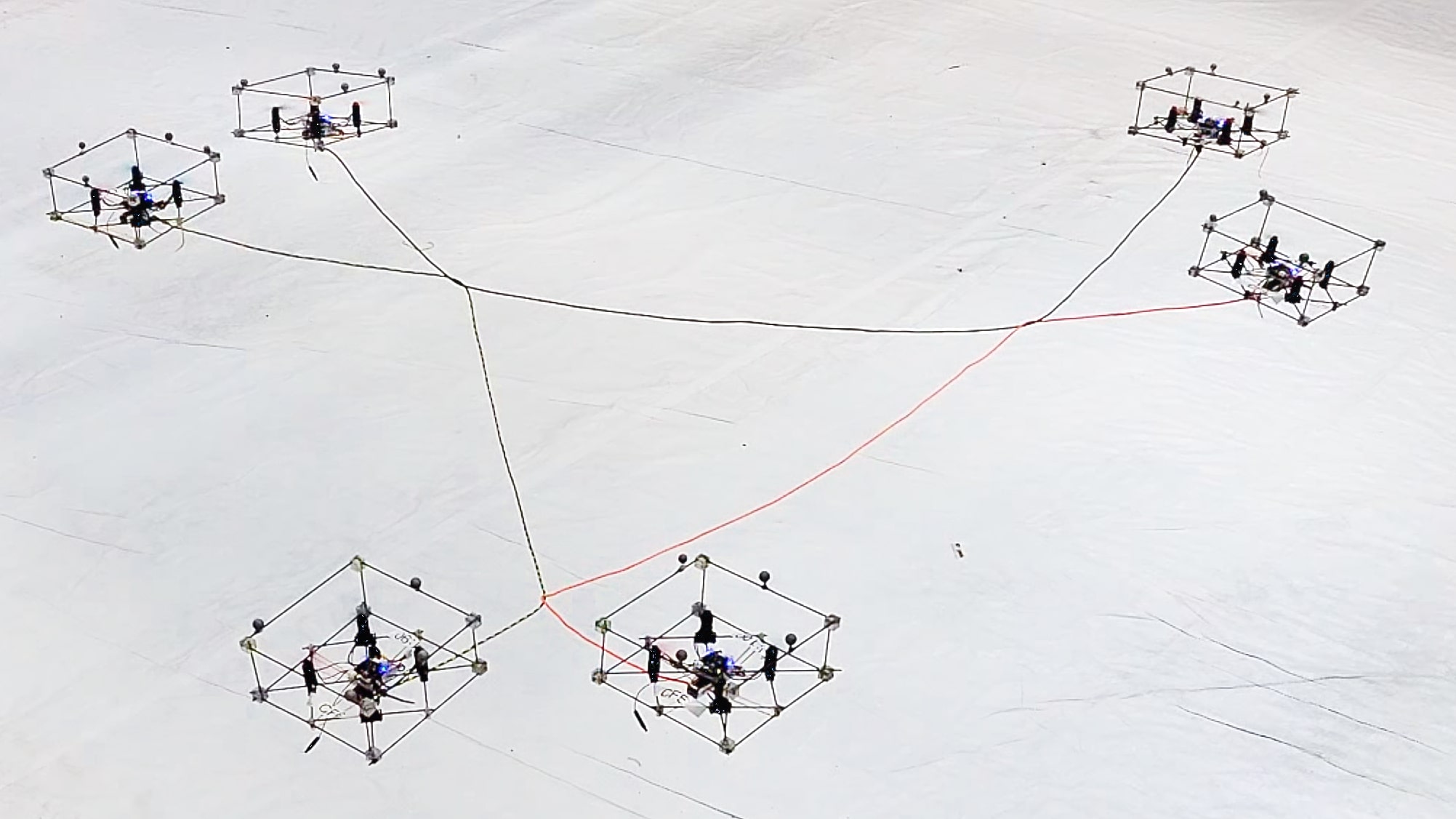}
  \caption{Triangular hitch}
  \label{fig:exp_braid_c}
\end{subfigure}

%
\begin{subfigure}{0.75\linewidth}
  \centering
  \includegraphics[width=\textwidth]{figures/snapshot13-min.jpg}
  \caption{Square hitch}
  \label{fig:exp_braid_f}

\end{subfigure}
    \caption{Polygonal hitches.}
    \label{fig:exp_braid}
\end{figure}

\textit{\textbf{Experiment 1 -  Action 1, forming a hitch:}} 
We propose a method to form a hitch by interlacing \correction{cables} in mid-air without human intervention. 
For a given polygon $\mathcal{P}$, we compute a balanced configuration following the procedure in Section~\ref{sec:robots_positions}, including the desired quadrotor positions $\mathbf{q}_k$ and $\mathbf{r}_k$ for $k=1,\dots,n$.
Then, we follow the algorithm in Section \ref{subsec:forming a hitch} for a polygon $\mathcal{P}~=~\{\mathbf{p}_1,\mathbf{p}_2,\mathbf{p}_3\}$,  where $\mathbf{p}_1=[0., -1., \correction{0.5}]^\top$, $\mathbf{p}_2=[-0.9, 0.4, \correction{0.5}]^\top$, and $\mathbf{p}_3=[0.9, 0.4, \correction{0.5}]^\top$.
%
%
In Step 1, the quadrotors move to the starting point as illustrated in Fig. \ref{fig:exp_braid_a}. 
In Step 2, they swap position with a circular trajectory between $\mathbf{q}_k$ and $\mathbf{r}_k$ creating an intersection  (see Fig. \ref{fig:exp_braid_b} and \ref{fig:exp_braid_c}). 
In Step 3, the robots  complete the polygonal-hitch (see Fig. \ref{fig:exp_braid_d} and \ref{fig:exp_braid_e}). 
We performed the experiment multiple times and found that a convex \SB{regular} polygon was a reliable polygon to form a hitch. Otherwise, if the convex polygon has a wide angle between the edges $(\mathbf{p}_k, \mathbf{p}_{k+1})$ and $(\mathbf{p}_k, \mathbf{p}_{k-1}$, the swapping trajectory has a bigger radius. This is easy to check, because the radius is half of the \correction{Euclidean} distance between $\mathbf{p}_k$, and $\mathbf{q}_{k})$.
%
%
Here we found that our success \SB{rate for the hitch forming} experiments \SB{is} 8 out of 10. The downwash can affect the robots' trajectory during the location swapping step. However, it could be improved using a trajectory that maintains a higher vertical distance between \SB{the} robots \cite{James2017}.
We also \SB{successfully} formed a hitch with four vertices as shown in Fig. \ref{fig:exp_braid_f}.

%



\begin{figure}[t]
    \centering
    \includegraphics[width=0.45\textwidth]{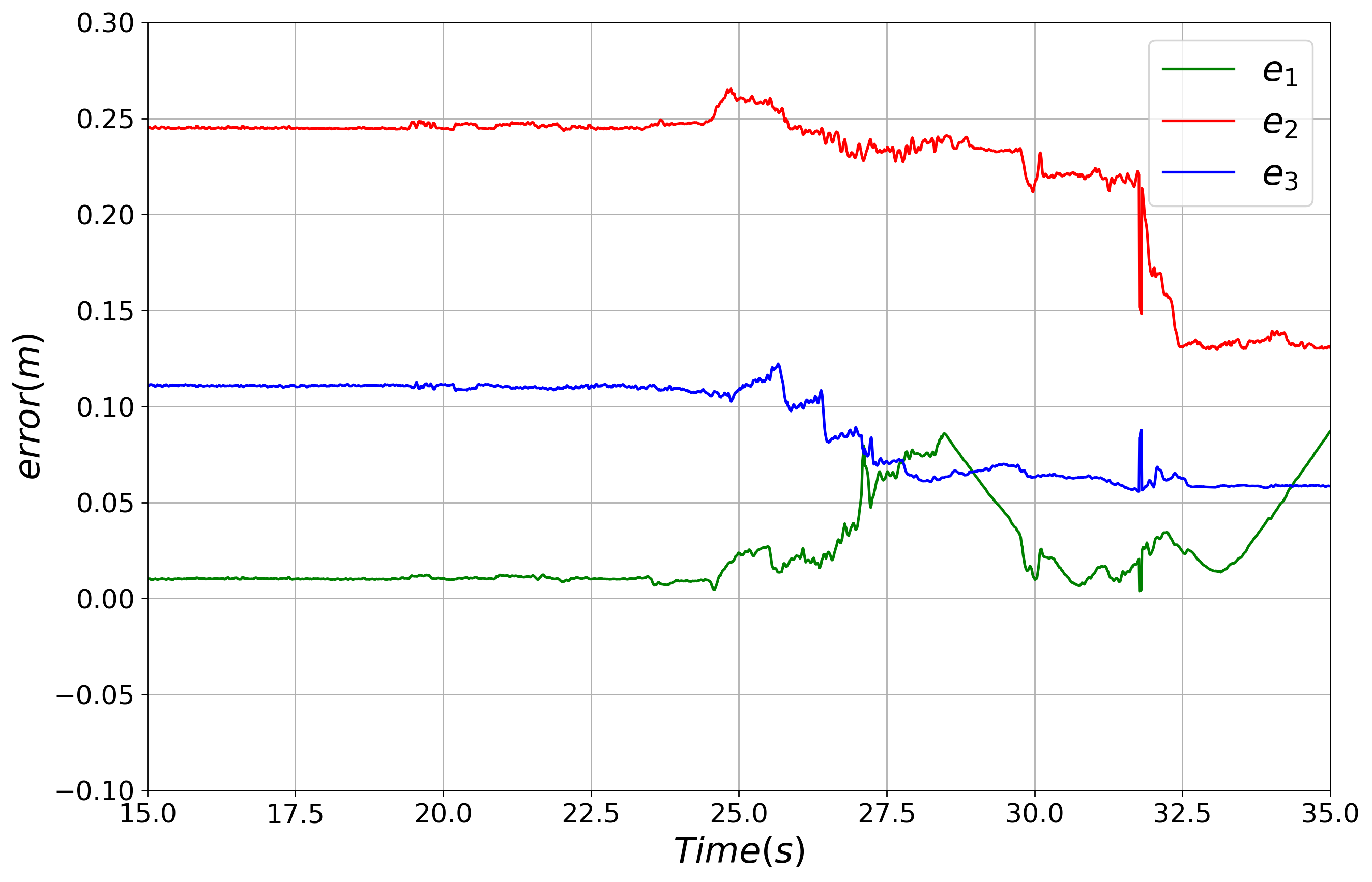}
    \caption{Results of Experiment 2: Action moving a vertex. The plots are showing the error between the desired intersection and the actual intersection.
    \correction{The green line} shows the error for the \correction{moving point $\mathbf{p}_1$}. \correction{Red and blue lines show the error of fixed points $\mathbf{p}_2$ and $\mathbf{p}_3$ respectively.}} 
    \label{fig:experiment2}
\end{figure}

\textit{\textbf{Experiment 2 - Action 2, moving a vertex:} }
This action is focused on changing the shape of the polygon by moving a single vertex. 
To demonstrate that a polygon\SB{al}  hitch is able to control \SB{an} intersection point,
we perform an experiment where the point $\mathbf{p}_1$ is moving along a trajectory described by,
\begin{equation*}
    \mathbf{p}_1(t) = \left \{
    \begin{array}{cc}
        (0, -0.1) &\text{ if }t < 25s \\
        (0, 0.05t - 0.1) &\text{ if }t > 25s
    \end{array}
    \right.
\end{equation*}
The  input is the \SB{initial} polygon
 $\mathcal{P}~=~\{\mathbf{p}_1,\mathbf{p}_2,\mathbf{p}_3\}$,  where $\mathbf{p}_1=[0., -1., \correction{0.5}]^\top$, $\mathbf{p}_2=[-0.4, 0.5, \correction{0.5}]^\top$, and $\mathbf{p}_3=[0.4, 0.5, \correction{0.5}]^\top$. 
We compute the Euclidean distance between the current intersection points and the desired intersection points -- see results in Fig. \ref{fig:experiment2}. The average error in position \SB{are} $\mu_{p_1} = 0.0819$, $\mu_{p_2} = 0.168$, and $\mu_{p_3} = 0.762$, and its standard deviations \SB{are} $\sigma_{p_1} = 0.056$, $\sigma_{p_2} = 0.145$, and $\sigma_{p_3} = 0.088$.

\correction{Since} our current implementation does not use any feedback from the actual position of the intersection points, \correction{there are some} offsets in their positions. \correction{It is} interestingly, \correction{that} the \correction{moving} point $\mathbf{p}_1$ has \correction{a smaller error.} \correction{We observed that} the motion of the vertex \correction{helps to} overcome \correction{the} frictional resistance \correction{between cables}.
We tested two types of ropes -- nylon and leather. The nylon rope has \correction{higher} friction, and it makes \SB{it more} difficult to move the vertex. 



\begin{figure}[t]
    \centering
    \includegraphics[width=0.45\textwidth]{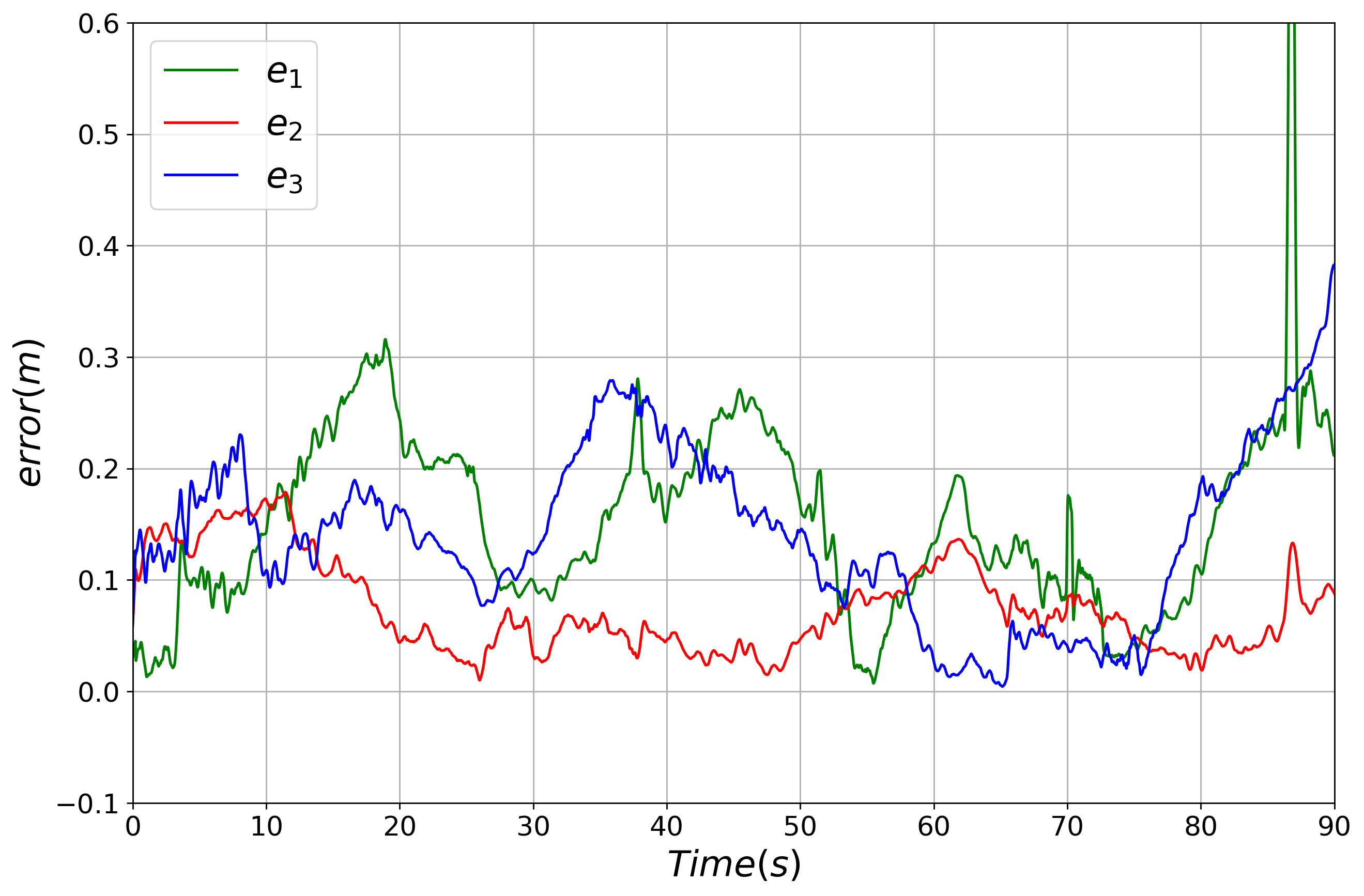}
    \caption{Results of Experiment 3: Action moving an edge. The plots show the error of the desired intersection.
    \correction{The static point} $\mathbf{p}_1$ \correction{is represented by the} \correction{green line}. 
    \correction{The moving points $\mathbf{p}_2$ and $\mathbf{p}_3$ are the red and blue line, respectively.} }
    \label{fig:experiment3}
\end{figure}

\textit{\textbf{Experiment 3 - Action 3, moving an edge,} }
This experiment shows the ability to move an edge. 
%
The initial 
polygon, 
$\mathcal{P}~=~\{\mathbf{p}_1,\mathbf{p}_2,\mathbf{p}_3\}$, 
\SB{is given by}
 $\mathbf{p}_1=[0., -1., \correction{0.5}]^\top$, $\mathbf{p}_2=[-0.4, 0.5, \correction{0.5}]^\top$, and $\mathbf{p}_3=[0.4, 0.5, \correction{0.5}]^\top$. 
We computed the distance between the current intersection points and the desired intersection points (see results in Fig. \ref{fig:experiment3}). \correction{However, we observed an increase in position errors in $e_1$ while $e_2$ and $e_3$ were moving in the second ten. This increase in error is due to the friction of the interlaced cable.}  The average error in positions \SB{are} $\mu_{p_1} = 0.0819$, $\mu_{p_2} = 0.168$, and $\mu_{p_3} = 0.762$, and \SB{the} standard deviations \SB{are} $\sigma_{p_1} = 0.056$, $\sigma_{p_2} = 0.145$, and $\sigma_{p_3} = 0.088$.

\begin{figure}[t]
    \centering
    \includegraphics[width=0.45\textwidth]{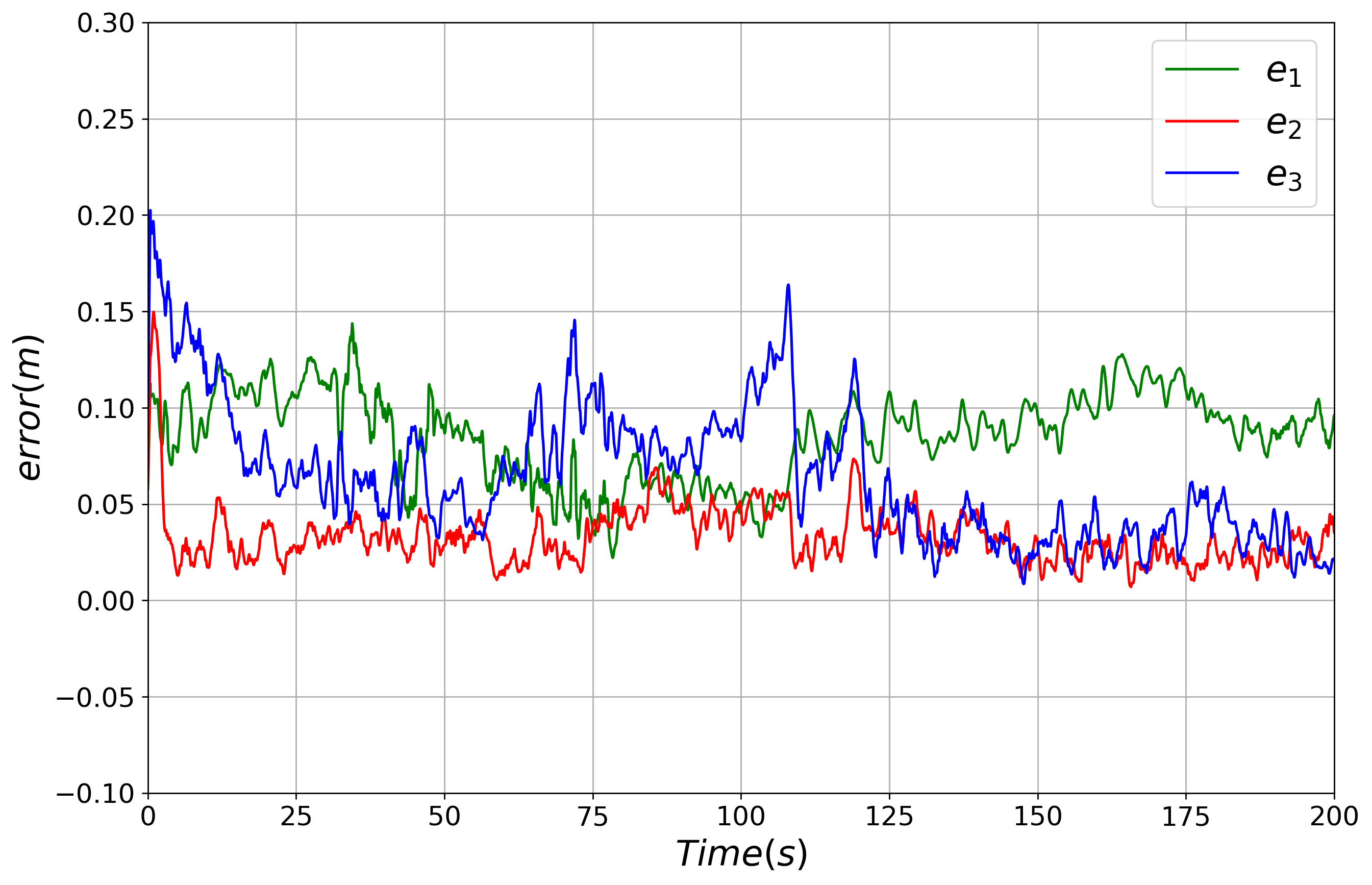}
    \caption{
    Results of Experiment 4: Action adjusting the cable. The plots show the error of the desired intersection.
    $\mathbf{p}_1$, $\mathbf{p}_2$ and $\mathbf{p}_3$ .}
    \label{fig:experiment4}
\end{figure}

\textit{\textbf{Experiment 4 - Action 4, adjusting a cable:} }
The last action that we discuss is adjusting the cable length. This action is helpful after moving a vertex or an edge as the resultant action will give $e_k \not= d_k$. Then, adjusting the cable length \SB{can create a balanced configuration with} $e_k=d_k$.
We computed the distance between the current intersection points and the desired intersection points (see results in Fig.~\ref{fig:experiment4}). 
\correction{Notably, this experiment shows the ability to adjust the cable and maintain a smaller position error.}
The average error in position $\mu_{p_1} = 0.0363$, $\mu_{p_2} = 0.026$, and $\mu_{p_3} = 0.043$, and its standard deviations $\sigma_{p_1} = 0.036$, $\sigma_{p_2} = 0.084$, and $\sigma_{p_3}~=~0.987$.

%



\section{Conclusion and future work}
\label{section: conclusions}

 In this work, we propose a novel \correction{class} of hitch\correction{es} that can be formed and \correction{morphed} in midair using a team of aerial robots with cables.
We introduce the concept of a \emph{polygonal-hitch}, which consists of multiple catenary robots forming a cyclic sequence by interlacing multiple cables.
The cables of two consecutive catenary robots are linked, forming a convex polygonal shape. 
We propose an algorithm to form the hitch systematically without any human intervention. The steps can run in parallel,
allowing hitches with a large number of robots to be formed in constant time.
We develop a set of actions that include \correction{three actions} to change\correction{its} shape.
%
Including 
, moving a vertex, moving an edge, and adjusting the cable.
We analyzed and controlled the hitch with a quasi-static approach that works in simulation and actual robots. 
%
We demonstrated the successful functionality
of our system in simulation and actual robots. In future work, we aim to transport objects and include the cable dynamics involved in our system.
%

%
\bibliographystyle{ieeetr}
\bibliography{referencias.bib}

\end{document}